\documentclass[conference]{IEEEtran}
\usepackage{cite}

\ifCLASSINFOpdf
  \usepackage[pdftex]{graphicx}
  \graphicspath{{../pdf/}{../jpeg/}}
  \DeclareGraphicsExtensions{.pdf,.jpeg,.png}
\else
\fi
\usepackage{amsmath}
\usepackage{array}

\usepackage{courier}
\usepackage{tikz}
\def\checkmark{\tikz\fill[scale=0.4](0,.35) -- (.25,0) -- (1,.7) -- (.25,.15) -- cycle;} 
\usepackage{multirow}

\begin{document}
%
\title{A Comparison of Deep Learning Object Detection Models for Satellite Imagery}

\author{\IEEEauthorblockN{Austen Groener}
\IEEEauthorblockA{Lockheed Martin Space\\
King of Prussia, Pennsylvania\\
Austen.M.Groener@lmco.com}
\and
\IEEEauthorblockN{Gary Chern}
\IEEEauthorblockA{Lockheed Martin Space\\
Palo Alto, California\\
Gary.Chern@lmco.com}
\and
\IEEEauthorblockN{Mark Pritt}
\IEEEauthorblockA{Lockheed Martin Space\\
Gaithersburg, Maryland\\
Mark.Pritt@lmco.com}}


\maketitle

\begin{abstract}
In this work, we compare the detection accuracy and speed of several state-of-the-art models for the task of detecting oil and gas fracking wells and small cars in commercial electro-optical satellite imagery. Several models are studied from the single-stage, two-stage, and multi-stage object detection families of techniques. For the detection of fracking well pads (50m-250m), we find single-stage detectors provide superior prediction speed while also matching detection performance of their two- and multi-stage counterparts. However, for detecting small cars, two-stage and multi-stage models provide substantially higher accuracies at the cost of some speed. We also measure timing results of the sliding window object detection algorithm to provide a baseline for comparison. Some of these models have been incorporated into the Lockheed Martin Globally-Scalable Automated Target Recognition (GATR) framework.
\end{abstract}


%
\IEEEpeerreviewmaketitle

\section{Introduction}

\begin{figure}[htb]
\centering
\includegraphics[width=3.4in]{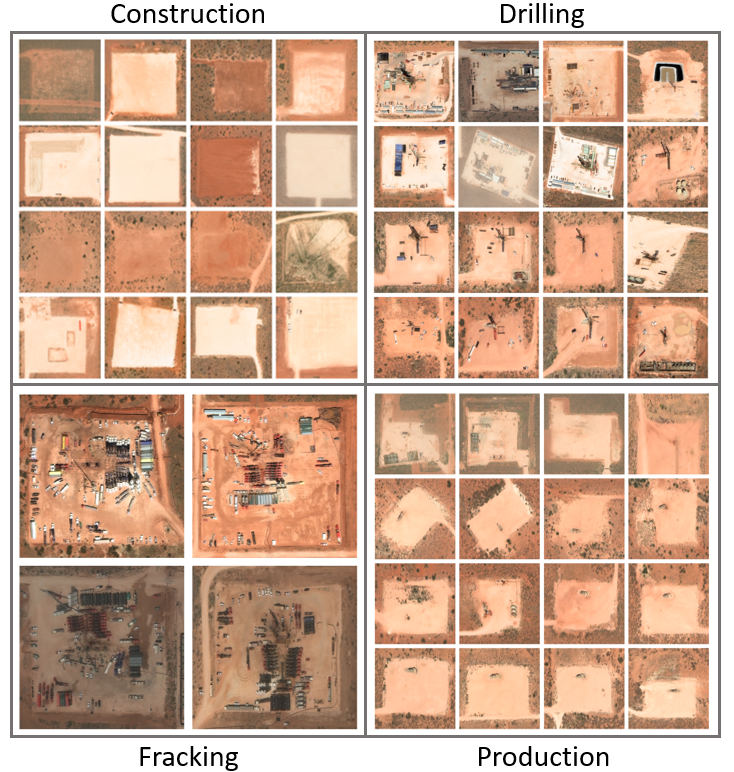}
\caption{Examples of fracking wells in various phases in New Mexico, USA. Images: DigitalGlobe, Inc}
\label{nm_wells}
\end{figure}

Both the image analyst and the Earth scientist face new and escalating challenges relating to the extraction of intelligence from the volume of data collected by multi-spectral, overhead sensors. Automated target recognition (ATR)\textemdash also referred to as object detection\textemdash will play a central role in automating and augmenting tasks such as broad area search, persistent site monitoring, discovery and tracking of construction sites, land management analysis, humanitarian aid and disaster relief, and monitoring oil and gas fracking wells (Figure \ref{nm_wells}). The output of such object detection algorithms\textemdash the target class and bounding box coordinates\textemdash are then fed into numerous downstream applications and analytics.

Deep learning models, more specifically convolutional neural networks (CNN), are increasingly being used as the core enabling technology for detecting and classifying objects within electro-optical (EO) \cite{8457969, yolt} and synthetic aperture radar (SAR) imagery \cite{8707419}. Due to the simplicity of its implementation, the sliding window algorithm is often used in place of more modern approaches for object detection found in the literature. However, sliding window suffers from a fundamental trade-off between speed and localization accuracy \cite{8707415}.

The highest accuracy models have almost always employed a two-stage approach to object detection, which can be broken down into the detection of candidate objects (Stage 1) followed by object classification and location refinement using a CNN (Stage 2). Single-stage models perform object detection in a single shot, accomplishing this by performing regression on the bounding box parameters along with classification simultaneously using a single network. Recently, several single-stage models have now shown to match performance of their two-stage counterparts while also providing significant speed improvements \cite{8237586, Li2018GradientHS}. Lastly, multi-stage models typically package several techniques together\textemdash whether it be of similar or disparate types\textemdash to accomplish object detection.

In this work, we compare the detection accuracy and speed measurements of several state-of-the-art models\textemdash RetinaNet \cite{8237586}, GHM \cite{Li2018GradientHS}, Faster R-CNN \cite{Ren2015FasterRT}, Grid R-CNN \cite{Lu2018GridR}, Double-Head R-CNN \cite{wu2019doublehead}, and Cascade R-CNN \cite{8578742}\textemdash for the task of object detection in commercial EO satellite imagery. To fully explore the solution space, we use ResNet-50\cite{He_2016}, ResNet-101 \cite{He_2016}, and ResNeXt-101 \cite{Xie2016AggregatedRT} CNN options as each model's backbone\textemdash to serve as the primary feature extraction mechanism. We then compare our timing results to the sliding window algorithm for baseline comparison. Some of these models have been incorporated into the Lockheed Martin Globally-Scalable Automated Target Recognition (GATR) framework \cite{GATR}.

\section{Problem}
The accurate and timely detection of objects within imagery produced by Earth observing satellites is a key technology leveraged by many sub-disciplines within the field of Earth science. As the volume of data increases rapidly, so too will the need for automated detection algorithms to continue to improve in speed and accuracy. The necessity for real-time object detection and tracking within frames of high-resolution video streams originates from the requirement that autonomous vehicles be able to swiftly respond\textemdash on the order of milliseconds\textemdash to avoid deadly collision. Out of this need arose an incredibly active area of research and development\textemdash in both algorithms and hardware\textemdash primarily leveraging progress in deep learning to advance the field of ground-based machine perception. Rather than applying these techniques to individual frames of high-resolution video data, we treat each chip within a much larger satellite image as independent input data.

Object detection in satellite imagery presents several unique challenges which largely differ from those found in the field of autonomous vehicles. Objects within satellite images tend to be viewed from directly overhead, though with some variation due to off-nadir viewing angles. Conversely, autonomous vehicles observe objects\textemdash like people, vehicles, buildings, and traffic signs\textemdash strictly from terrestrial vantage points. Among the other challenges are individual image size and data volume. DigitalGlobe's WorldView-3 commercial satellite alone is capable of collecting up to 680,000 $\mathrm{km}^{2}$ of panchromatic (31cm) imagery per day. A single WorldView-3 panchromatic image is typically over 10 Gigapixels, which is several orders of magnitude larger than the typical image taken by ground-based systems. Objects of interest in satellite imagery also span the range of physical scales, from meters (several to tens of pixels) to hundreds of meters (tens to hundreds of pixels), and are often partly or completely obscured by clouds, and can also be affected by things like temperature, humidity, haze, and sun angle, to name a few. Lastly, on the extreme end, objects may only be several pixels in size, while also being tightly packed together. 

\begin{figure}[t!]
\centering
\includegraphics[width=3.4in]{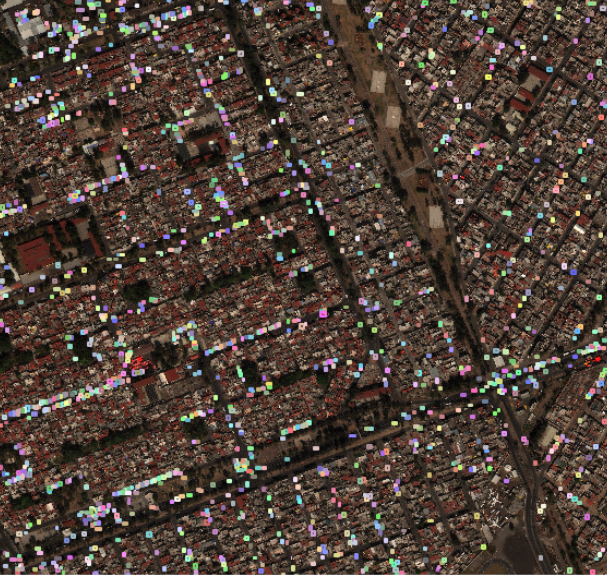}
\caption{Examples of densely packed small cars in an urban environment from the xView dataset. Each colored box represents one object our algorithms aim to detect.}
\label{xview_cars}
\end{figure}

\section{Dataset}
The main objective of this work is to explore the detection accuracy and inference speed of a number of modern object detection models, used primarily by the autonomous driving industry, and how they perform when reapplied to the task of detecting objects in overhead satellite imagery. We select two object types\textemdash oil and gas fracking wells and small cars\textemdash to represent detection on various physical scales.

\subsection{WorldView-3: Oil and Gas Fracking Wells}
Hydraulic fracturing, or ``fracking", is one of several types of unconventional oil and gas extraction techniques. This process involves drilling into shale formations (a type of sedimentary rock) followed by hydraulically fracturing the rock in order to liberate the oil and gas trapped within \cite{fracking}. Large well pads (50m - 250m) composed of crushed rock are constructed in order to support the heavy equipment needed in the drilling, fracking, and pumping phases of the well, and this is what our algorithms aim to detect. 

Large, accurately labeled datasets of fracking wells needed for training supervised deep learning models do not currently exist. We therefore manually curate over 12,000 fracking wells contained in WorldView-3 pansharpened imagery (0.31m GSD)\textemdash in all phases of operation\textemdash from around the globe in order to build a globally-scalable object detection model \cite{GATR}. In this work, however, we focus on a subset of the data in the state of New Mexico, constituting 2,802 examples of fracking wells contained in 568 5000x5000 pixel WorldView-3 image chips. We then down-sample further to resolutions of 1250x1250 (1.24m GSD) and 1050x1050 (1.48m GSD), when training and testing models with ResNet-50/101 and ResNeXt-101 backbones, respectively.

\subsection{xView: Small Cars}
The second object we focus on is the small car category (class \#18; Figure \ref{xview_cars}) from the xView dataset \cite{Lam2018xViewOI}. The xView dataset\footnote{“DIUx xView 2018 Detection Challenge,” Defense Innovation Unit Experimental (DIUx) and the National Geospatial-Intelligence Agency (NGA), https://xviewdataset.org} is a massive, publicly available satellite imagery dataset (along with hand-annotated ground-truth bounding boxes for 60 classes) organized by the Defense Innovation Unit Experimental (DIUx) and the National Geospatial-Intelligence Agency (NGA). The purpose of the 2018 xView challenge was to advance the field of computer vision and develop new solutions for national security and disaster response. Applying computer vision to overhead imagery has the potential to detect emerging natural disasters, improve response, quantify the direct and indirect impact\textemdash and save lives.

Within the dataset, there are 210,938 examples of small cars\textemdash roughly 35 percent of all images contain at least one car\textemdash with contextual environments ranging from dense urban scenes to sparsely populated country roads. At the native spatial resolution of the imagery (30cm GSD), small cars may only span perhaps 7 pixels on a side (median: 14 pixels on a side), and thus pose a real challenge for most out-of-the-box algorithms due to size and crowding. Figure \ref{xview_cars} shows one example of an image our algorithms will need to make predictions on. We leave the imagery at its native resolution. However, we partition each image into chips of 600x600 pixels and then up-sample to 1200x1200 and 1000x1000 when training and testing with ResNet-50/101 and ResNeXt-101 backbones, respectively. Annotations for small cars are typically 10x16 pixel in size at the native resolution of the Worldview-3 satellite, which is less than half the size of the smallest Faster R-CNN anchor of 32x32, for instance. This requires reducing the intersection over union (IoU) thresholds (see Section \ref{metrics}) for proper assignment of anchor boxes to ground truth, or upsampling the image by a factor of 2. We choose the latter method to allow us to use the default model configuration settings.

\begin{table}[t!]
\footnotesize
\centering
\begin{tabular}{ |p{2.48cm}||>{\centering\arraybackslash} p{1.5cm}|>{\centering\arraybackslash} p{1.5cm}|>{\centering\arraybackslash} p{1.55cm}|  }
 \hline
 \multicolumn{4}{|c|}{Detection Model + Backbone Combinations} \\
 \hline
 Model Architecture & ResNet-50 & ResNet-101 & ResNeXt-101  \\
 \hline
 RetinaNet & \checkmark & \checkmark & \checkmark \\ 
 GHM & \checkmark & \checkmark & \checkmark \\
 Faster R-CNN & \checkmark & \checkmark & \checkmark \\ 
 Grid R-CNN & \checkmark & \checkmark & \checkmark \\ 
 Double-Head R-CNN & \checkmark & \checkmark & \checkmark \\
 Cascade R-CNN & \checkmark & \checkmark & \checkmark \\ 
 \hline
 Sliding Window & \checkmark & \checkmark &  \\
 \hline
\end{tabular}
  \caption{Model and CNN backbone combinations used in this study.}
  \label{tab:models}
\end{table}

\section{Methods}
In this section, we describe details of the object detection models that were used in this study, followed by definitions of the metrics used to score them. For a complete summary of models, see Table \ref{tab:models}.

\subsection{State-of-the-Art Models}
To ensure we properly benchmark each model's performance, we utilize a library called \texttt{MMDetection} \cite{Chen2019MMDetectionOM}. \texttt{MMDetection} is an open-source object detection toolbox (created using the PyTorch framework) which provides pre-trained weights and model definitions for over 200 high-performing object detection models, along with tools to train and test them. Though there are a numerous architectures and techniques in use today, \texttt{MMDetection} standardizes deep learning object detection models by expressing single-stage, two-stage, and multi-stage detection models into common functional pieces. For instance, all models share a common component called a {\it backbone}, which is typically a deep convolutional neural network (CNN) used to extract features from the input image to be used in downstream bounding box regression and classification components. 

We select two single-stage detectors (RetinaNet and GHM), three two-stage detectors (Faster R-CNN, Grid R-CNN, and Double-Head R-CNN), and a single multi-stage detector (Cascade R-CNN) for training and testing on our objects of interest. Though each model was created to solve a particular problem associated with object detection\textemdash whether it be speed, classification, or localization accuracy\textemdash it should be noted that none were designed specifically for use with overhead imagery (EO, IR, or SAR). 
\newline \newline
{\bf RetinaNet:} Until the introduction of RetinaNet, single-stage object detection models historically lagged in performance compared to more complex architectures. This model combined several ideas from previous works, namely the concept of anchor boxes \cite{Ren2015FasterRT} and the use of feature pyramids \cite{Liu2016SSDSS, Lin2016FeaturePN}, with a novel loss function referred to as {\it Focal Loss}. Focal Loss was designed to address the severe foreground-background class imbalance which was discovered to be the cause of single-stage detector performance degradation. By modifying the cross-entropy loss function to focus on the more difficult training examples, RetinaNet achieved state-of-the-art performance.
\newline \newline
{\bf GHM:} With the notion that several training imbalances exist, for example between easy and hard examples, and between positive and negative examples, a gradient harmonizing mechanism (GHM) was explored which modifies the standard classification and regression loss functions, GHM-C and GHM-R, respectively. Coupled with model architectures like RetinaNet, it was shown to achieve state-of-the-art performance.
\newline \newline
{\bf Faster R-CNN:} Faster R-CNN is the latest in a series of two-stage detection algorithms which originally stems from the combination of region proposals with convolutional neural network feature extraction, {\it R-CNN} \cite{Girshick2013RichFH}. Faster R-CNN improves on its predecessor, Fast R-CNN \cite{fastrcnn}, by using a second neural network for region proposal, rather than using selective search.
\newline \newline
{\bf Grid R-CNN:} This two-stage model approach to object detection differs from others in one key area\textemdash the bounding box localization technique. While most models improve object detection in one way or another, most perform bounding box localization in a similar way. Rather than formulating localization as a regression problem performed by a set of fully connected layers which take in high-level feature maps to predict candidate boxes, Grid R-CNN replaces this approach by utilizing a grid point guided localization mechanism, and demonstrates state-of-the-art performance on the MS COCO \cite{Lin2014MicrosoftCC} benchmark, especially when tightening up the object localization constraint. Microsoft Common Objects in Context (COCO) is a large-scale object detection, segmentation, and captioning dataset used for benchmarking algorithms.
\newline \newline
{\bf Double-Head R-CNN:} Models based on R-CNN tend to apply a single head to extract regions of interest (RoI) features for both classification and localization tasks. Instead, the Double-Head R-CNN uses a fully-connected head for classification while using a convolutional head for bounding box regression. With this modification, they were able to achieve gains over previous architectures on the MS COCO dataset. 
\newline \newline
{\bf Cascade R-CNN:} Cascade R-CNN stands as the only multi-stage detection model we explore in this work. This model builds off of R-CNN by daisy-chaining a series of detectors together\textemdash each more strict about object localization than the last\textemdash in order to be sequentially more selective against close false positives. The detectors are trained
together stage by stage, with the notion that the output of one detector is a good distribution for training the subsequent detector in the series.

\begin{figure}[!t]
\centering
\includegraphics[width=3.25in]{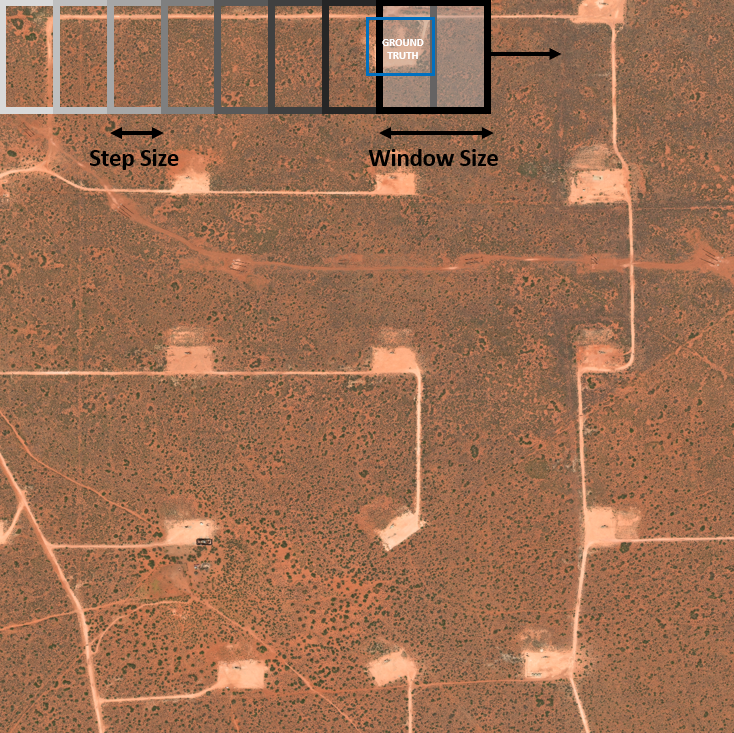}
\caption{Schematic of the sliding window object detection technique. Blue box represents a notional ground truth bounding box. Image: DigitalGlobe, Inc}
\label{swschematic}
\end{figure}

\subsection{Sliding Window Model}
We also benchmark the Sliding Window detector as a basis for comparison against the models selected above. Sliding window represents perhaps the simplest and most intuitive object detection system one might implement, leveraging a single CNN model which performs classification on a predefined window within the image data array. This window is then slid across the input image in both dimensions to achieve global detection coverage. Figure \ref{swschematic} exhibits several steps of this technique. 

The sliding window algorithm can be a very effective technique for ATR with a few caveats. Often at first, the perception is that sliding window is an easier technique to implement. However, special care must be taken to select the right window size for the target under consideration. Furthermore, sliding window does poorly in the event that object sizes vary drastically, and if several objects (at various physical sizes) are considered, then often several sliding window models must be trained and used together. If any of these conditions are not met, the performance suffers.

Standalone ResNet-50 and ResNet-101 CNN classifiers are trained on down-sampled WorldView-3 image chips (1.24m GSD) of size 250x250, comprising a balanced dataset of fracking wells and clutter (defined as everything that is not a well). The window size is chosen to be the smallest size needed to completely cover all truth bounding box annotations within the combined training and validation sets, readjusted after downsampling from native resolution imagery. Due to the very high classification accuracies on the validation set for both ResNet-50/101 models, it is deemed unnecessary to train the more complex ResNeXt-101 model.

\subsection{Training Details}
The MM-Detection library provides default training configuration files and pre-trained ImageNet weights for each model architecture, which we built off of for each CNN backbone chosen. Generally, most hyperparameter settings (except for learning rate) were left in their default setting. A Stochastic Gradient Descent (SGD) optimizer with momentum, learning rate decay, and a learning rate schedule was used in the training of all models. For fracking wells, the number of training epochs varied between 12 and 25 depending on model complexity, while we varied the number of epochs between 6 and 12 for cars. {\it Data augmentation} is an approach which increases the diversity of data available for training large models, without actually having to collect new data. We use random vertical and horizontal flips as data augmentation steps during model training.

Sliding Window CNN backbones (ResNet-50 and ResNet-101) are trained as standalone classifiers using target image chips rather than full scenes. Augmentation steps and optimization settings are nearly identical to the above models, with the exception of learning rate and the number of training epochs.

\subsection{Evaluation Metrics} \label{metrics}
We choose the \textbf{average precision} (AP), \textbf{max $\mathrm{F_{1}}$ score}, and \textbf{area coverage rate} metrics to score our object detection models. The prediction outputs of each model are the bounding box locations and associated class probabilities for each input image. Consequently, a decision must be made in order to concretely define what constitutes a detection in terms of probability and box localization accuracy. These parameters are the \textbf{probability score threshold} and \textbf{Jaccard Index} (or intersection over union, \textbf{IoU}). Given two bounding boxes, $B_{1}$ and $B_{2}$, the IoU is defined as
\begin{equation}
    IoU = \frac{\left| B_{1} \cap B_{2} \right|}{\left| B_{1} \right| + \left| B_{2} \right|  - \left| B_{1} \cap B_{2} \right|}
\end{equation}

With some particular choice of probability score and IoU thresholds, the precision and recall are defined as follows
\begin{equation}
    P = \frac{TP}{TP+FP}
\end{equation}
\begin{equation}
    R = \frac{TP}{TP+FN}
\end{equation}
where TP, FP, and FN represent the number of true positives, false positives, and false negatives, respectively.

The average precision is defined as the area under the precision-recall curve\textemdash formed by plotting precision vs recall as one sweeps over all possible probability scores. 
\begin{equation}
    AP = \int_{0}^{1} P(R) dR
\end{equation}
 In the case of multi-class object detection performance, the \textbf{mean average precision} (mAP) is typically used, which is simply the average of each class-specific AP score. The IoU which we choose to report the AP with is 0.50 (AP@IoU=0.50) for fracking wells\footnote{Oil and gas fracking well models incorporated into the GATR framework use a lower IoU threshold, leading to higher performance than what we will show in this work.}, and 0.25 (AP@IoU=0.25) for cars.

The max $\mathrm{F_{1}}$ score is the point on the precision-recall curve with the largest harmonic mean of the precision and recall. This roughly corresponds to being the point closest to the perfect classifier (P=1, R=1).
\begin{equation}
F_{1} = \frac{2 P R}{P + R}
\end{equation}

Finally, the area coverage rate measures the amount of data each model can process, and is measured in square kilometers per second.

\section{Results}
Timing results are collected using a single NVIDIA TITAN Xp GPU. The area coverage rate, measured in $\mathrm{km}^{2} / \mathrm{sec.}$, is obtained by measuring the run-time speed of the models in frames per second (FPS) on non-overlapping images of a fixed area. Regardless of the backbone network used (which dictates the amount of down-sampling performed) each WorldView-3 image used for the oil and gas fracking wells were each 2.4025 $\mathrm{km}^{2}$. Each image used for the xView small cars was 0.0324 $\mathrm{km}^{2}$. It should be noted that timing results for the sliding window detector are obtained with overlapping windows into the image data, which is precisely how sliding window achieves its object localization (and primarily why it is slower). While multiple GPUs were used to accelerate training of all models, we intentionally chose not to explore prediction-time parallelization techniques in this study.

\begin{figure}[htb]
\centering
    \includegraphics[width=\linewidth]{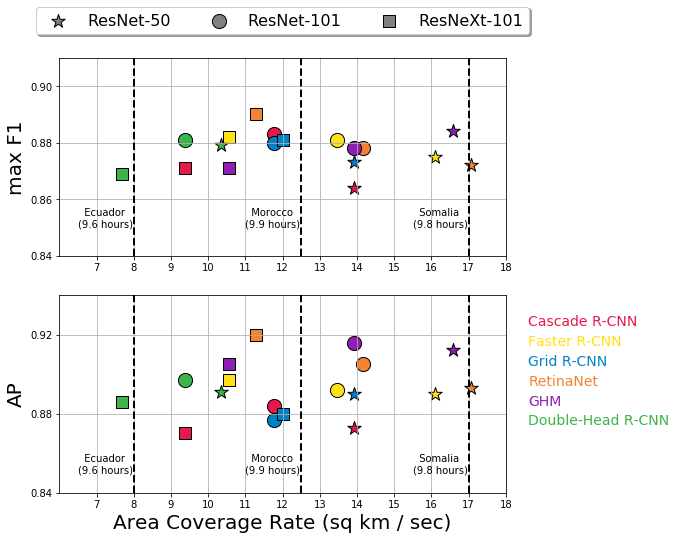}
    \caption{{\bf Oil and Gas Fracking Wells: }Speed and performance results for oil and gas fracking wells.}
    \label{allspeedswells}
\end{figure}

\subsection{Oil and Gas Fracking Wells}
 Overall, and somewhat surprisingly, the single-stage detectors RetinaNet and GHM provide the highest overall detection performance, while also providing the best area coverage rates. If highest possible performance is desired, RetinaNet with a ResNeXt-101 backbone is the top choice with an average precision of 0.92. If one optimizes strictly for area coverage rate, then RetinaNet with a significantly smaller backbone (ResNet-50) provides the best value. If one optimizes performance and timing jointly, for a minimal drop in max $\mathrm{F_{1}}$ and AP, GHM with ResNet-50 and ResNet-101 backbones, along with RetinaNet with a ResNet-101 backbone fill out the desirable portions of the trade space. For a visualization of the entire trade space, see Figure \ref{allspeedswells}.
 
\begin{figure*}[htb!]
\begin{tabular}{cc}
  \includegraphics[width=0.5\textwidth]{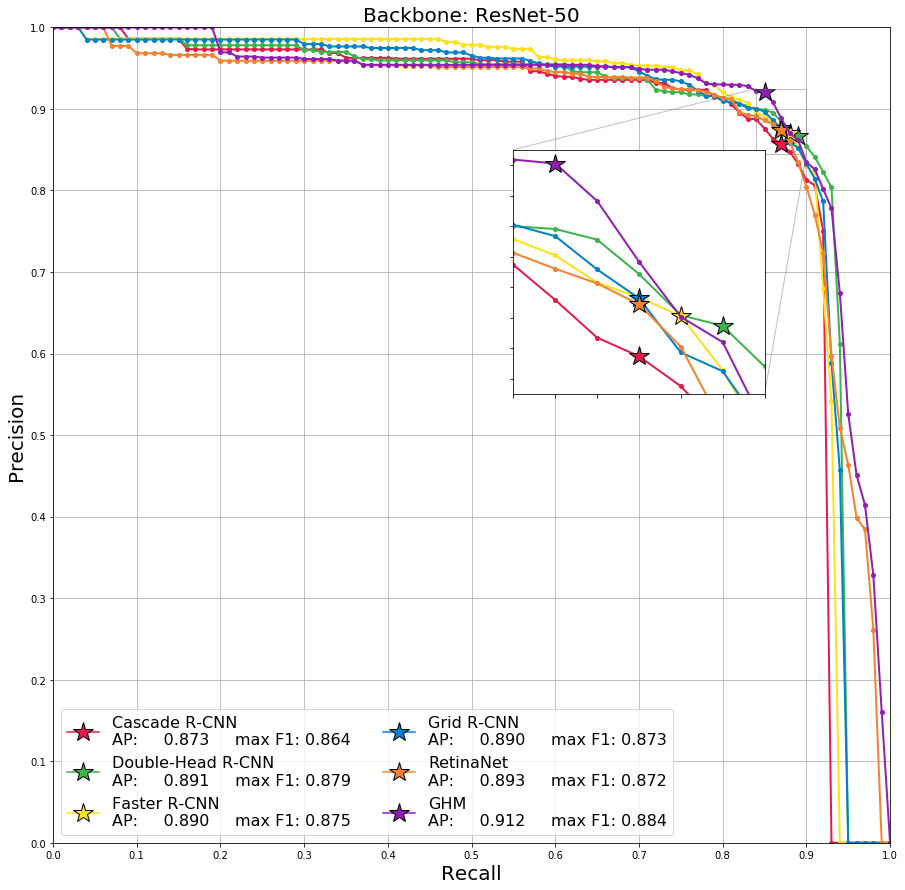} &   \includegraphics[width=0.5\textwidth]{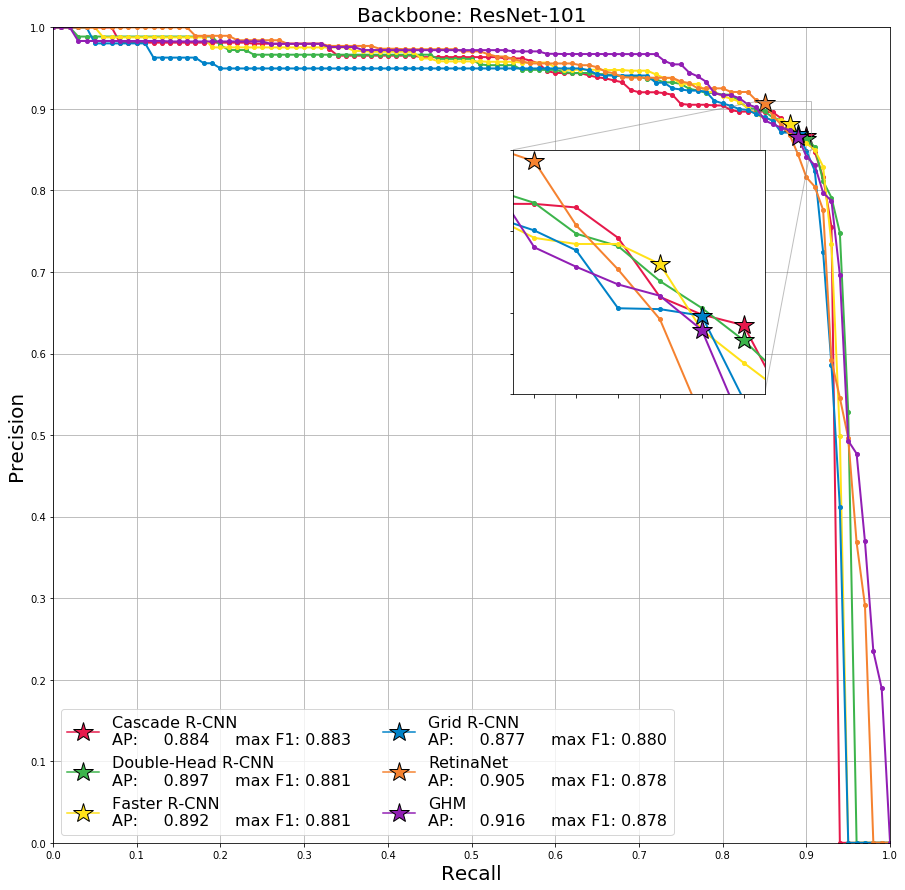} \\
(a)&(b)\\[6pt]
\multicolumn{2}{c}{\includegraphics[width=0.5\textwidth]{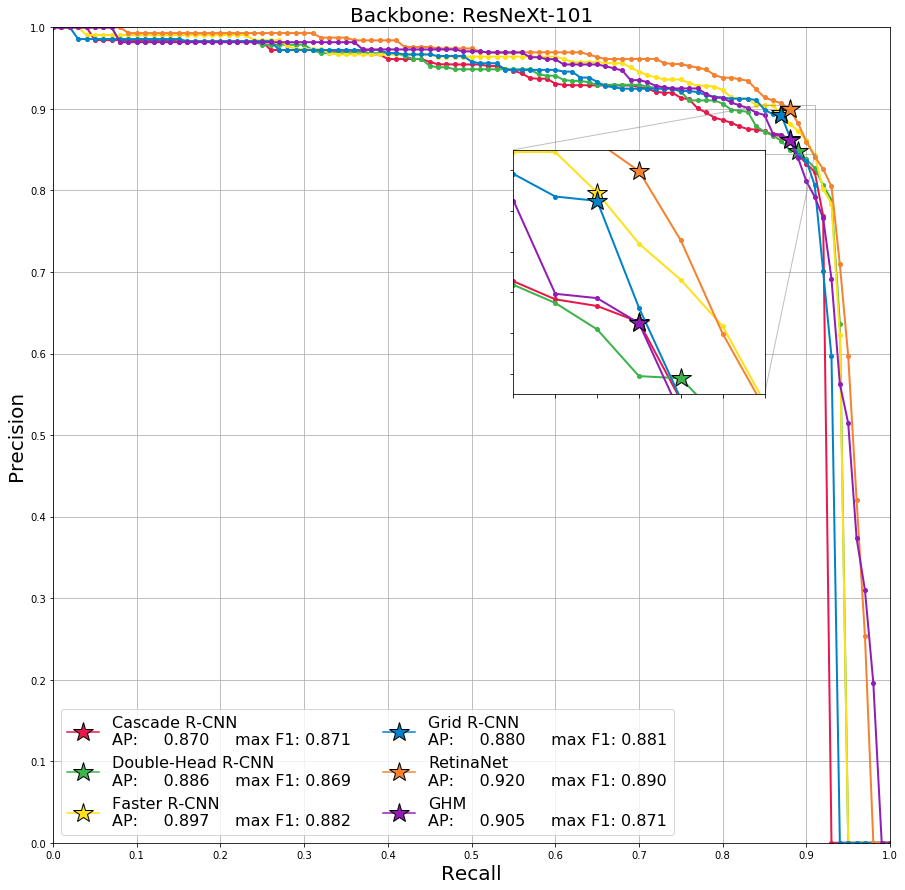} }\\
\multicolumn{2}{c}{(c)}
\end{tabular}
\caption{{\bf Oil and Gas Fracking Wells: }Precision-Recall (PR) curves for all object detection models for the following CNN backbones: (a) ResNet-50, (b) ResNet-101, (c) ResNeXt-101. Stars indicate the max $\mathrm{F_{1}}$ score, while average precision (AP) scores reported represent the area under the curve.}
\label{wellprcurves}
\end{figure*}

 Figure \ref{wellprcurves} shows precision-recall curves for each model using ResNet-50, ResNet-101, and ResNeXt-101 feature extractors, trained for detecting oil and gas fracking wells. Table \ref{tab:well_table_results} collects max $\mathrm{F_{1}}$ and average precision performance metrics along with timing results. 

\subsection{Small Cars}
\begin{figure}[htb]
\centering
    \includegraphics[width=\linewidth]{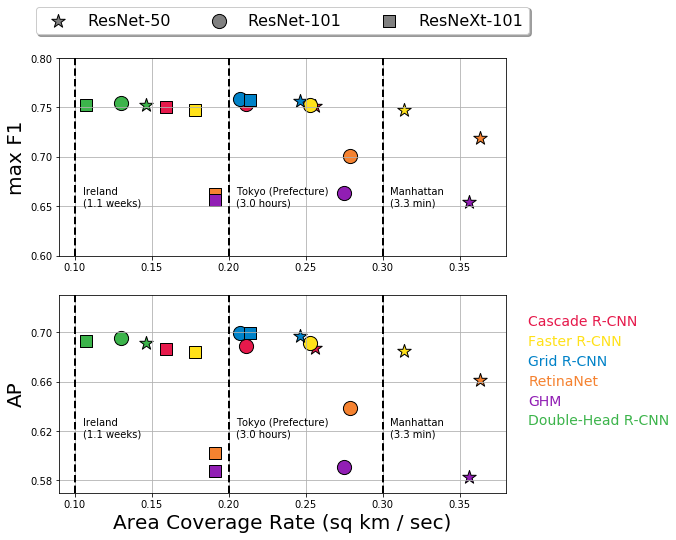}
    \caption{{\bf Small Cars: }Speed and performance results for small cars.}
    \label{allspeedscars}
\end{figure}

\begin{figure}[tb]
\centering
\includegraphics[width=\linewidth]{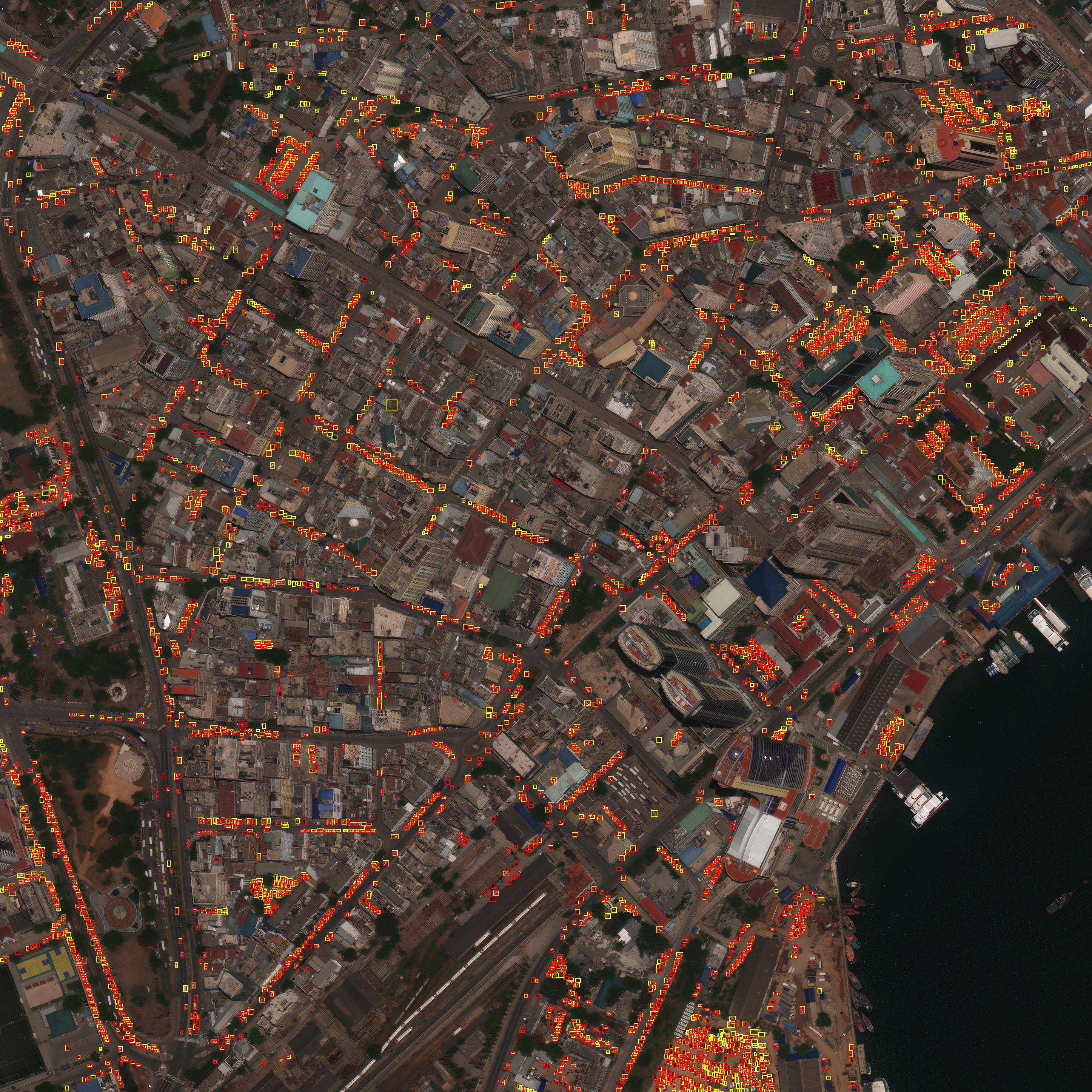}
\caption{Example prediction output from Grid R-CNN (with ResNet-101 backbone) on an xView validation image containing 3929 small cars. Yellow boxes are ground truth bounding boxes and red are predicted bounding boxes.}
\label{predexample}
\end{figure}

For the detection of cars, two-stage and multi-stage detectors all achieve average precisions of just under 0.70 and max $\mathrm{F_{1}}$ scores of around 0.75, while single-stage detectors lag in performance\textemdash varying in average precision from 0.661 (RetinaNet) down to 0.583 (GHM). The highest performing model is Grid R-CNN with ResNet-101 and ResNeXt-101 backbones, though several model and backbone combinations are very comparable. The speediest models are RetinaNet, GHM, and Faster R-CNN with ResNet-50 backbones. However, the best combination of speed and accuracy is Faster R-CNN with a ResNet-50 backbone. See Figure \ref{allspeedscars} for a visual comparison of all results. See also Figure \ref{predexample} for an example of predicted car locations compared to ground truth within a single xView validation image. It should be noted that performance metrics like these for vehicles in xView are typical due to size and crowding \cite{yolod}, and are expected to be lower than simpler objects like large fracking well pads. 

\begin{figure*}[htb!]
\begin{tabular}{cc}
  \includegraphics[width=0.5\textwidth]{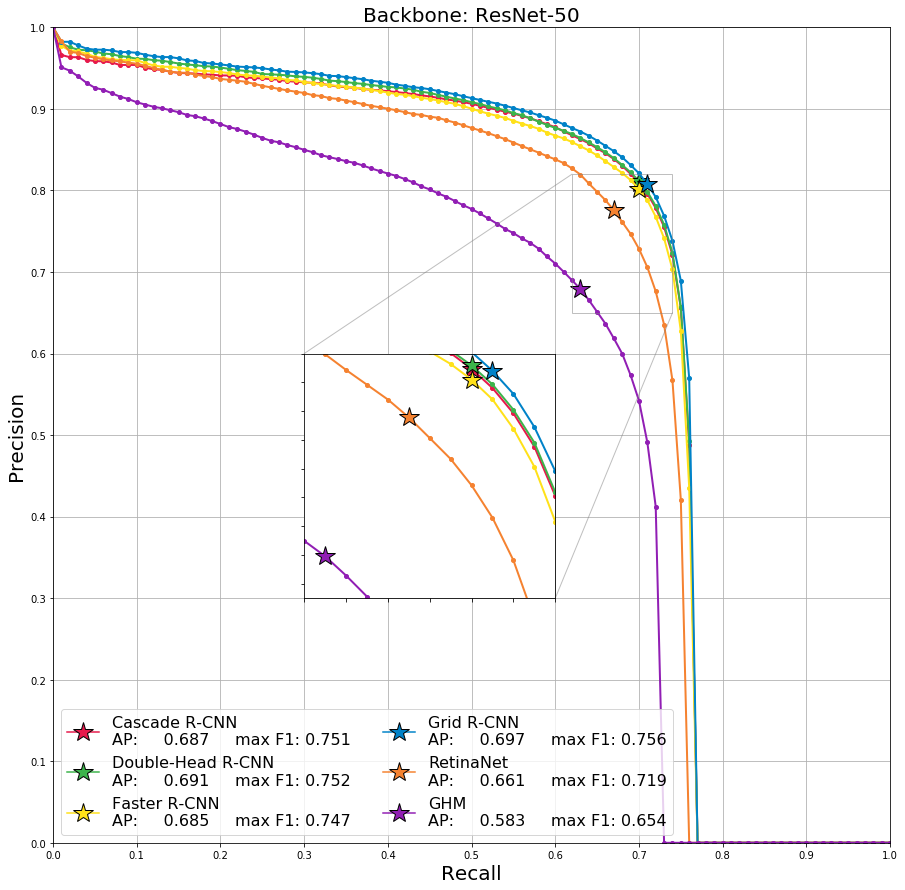} &   \includegraphics[width=0.5\textwidth]{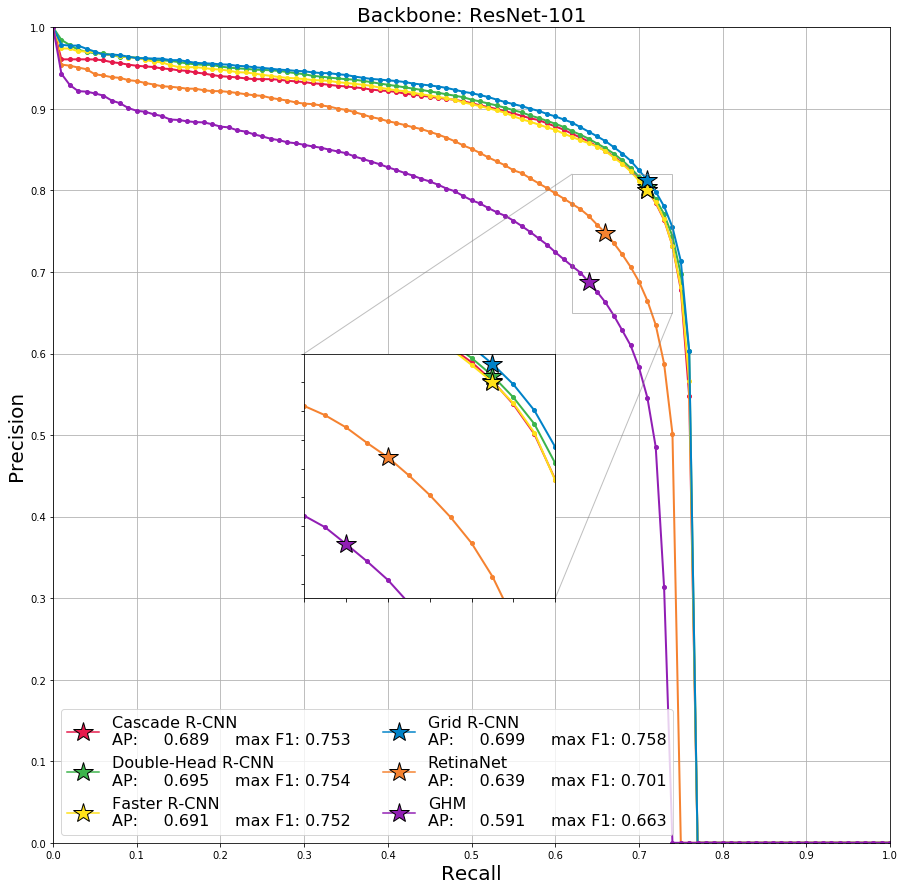} \\
(a)&(b)\\[6pt]
\multicolumn{2}{c}{\includegraphics[width=0.5\textwidth]{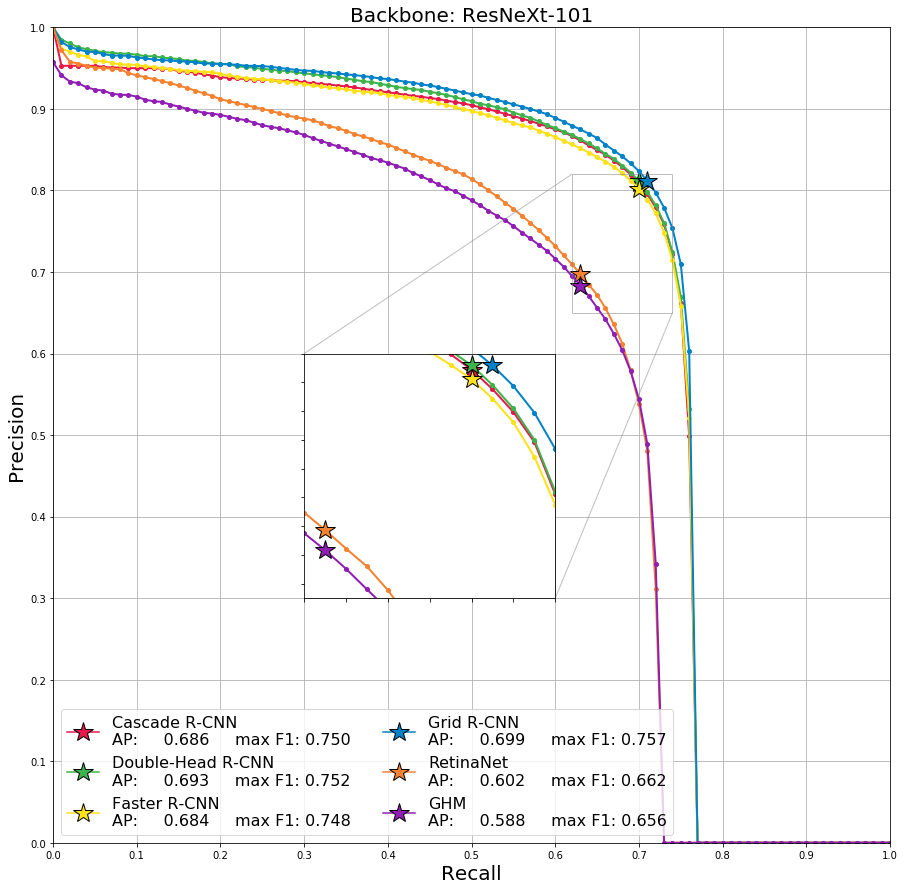} }\\
\multicolumn{2}{c}{(c)}
\end{tabular}
\caption{{\bf Small Cars: }Precision-Recall (PR) curves for all object detection models for the following CNN backbones: (a) ResNet-50, (b) ResNet-101, (c) ResNeXt-101. Stars indicate the max $\mathrm{F_{1}}$ score, while average precision (AP) scores reported represent the area under the curve.}
\label{carprcurves}
\end{figure*}

Figure \ref{carprcurves} shows precision-recall curves for each model using ResNet-50, ResNet-101, and ResNeXt-101 feature extractors, trained for detecting small cars. Table \ref{tab:car_table_results} collects max $\mathrm{F_{1}}$ and average precision performance metrics along with timing results.

\subsection{Sliding Window}
\begin{figure}[tb]
\centering
\includegraphics[width=3.25in]{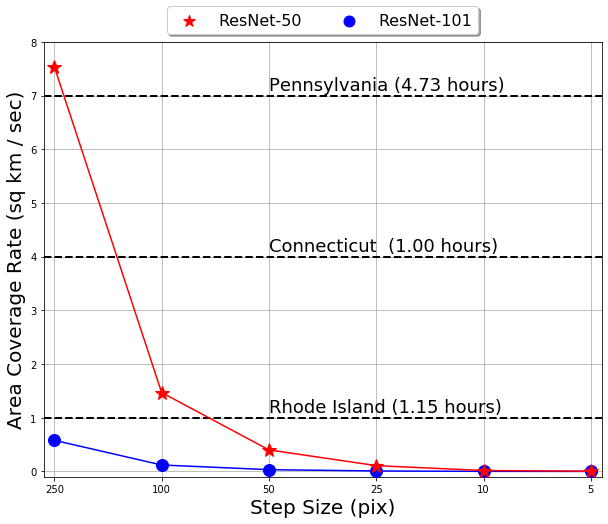}
\caption{Sliding window area coverage rates against step size for a window size of 250x250 (oil and gas fracking wells).}
\label{swspeeds}
\end{figure}

Figure \ref{swspeeds} highlights the trade-off between run-time and step size (and by proxy bounding box localization accuracy) one must make with the sliding window algorithm using ResNet-50 and ResNet-101 backbones. In order to achieve similar area coverage rates as modern object detection algorithms, impractically large step sizes (100 to 250 pixels) must be chosen.

\section{Conclusion}
In this work, we have conducted a study in which speed and performance of modern object detection algorithms were measured for the automatic detection of oil and gas fracking wells and small cars in commercial EO satellite imagery datasets. 

For the detection of wells, nearly all models perform satisfactorily, achieving both high marks in max $\mathrm{F_{1}}$ and average precision scores, though single-stage models seem to outperform their two-stage and multi-stage counterparts in both performance and speed. However, two-stage and multi-stage models significantly outperform single-stage models for the detection of small cars. A possible explanation for sub-par single-stage performance could be due to the difference in the amount of training data used for each object. The data support for xView small cars is larger by two orders of magnitude than the manually curated fracking well dataset. In this data-rich regime, single-stage models seem to lag in performance, while in the data-limited regime, they perform at similar levels compared to more complex models. Alternatively, a second possibility could be that the custom loss functions of RetinaNet and GHM require additional hyperparamter tuning for small objects such as cars.

In general, we find that the choice of CNN backbone used within each model architecture has a weak impact on performance metrics for detecting targets like fracking wells and small cars. It is likely the case that even the least complex backbone (ResNet-50) is still capable of effectively extracting the correct features to pass along to the model's head component, especially since we only consider models which aim to detect one object at a time. For models looking to detect 60 classes or more simultaneously, the backbone choice likely has more of a direct impact on performance. The decision about which backbone to use, however, does have a large impact on inference speed. We find that models which utilize ResNet-50 are capable of churning through the largest amount of data, followed by ResNet-101 and ResNeXt-101, which also directly reflects the size of each model (i.e., tunable parameters). In all practical settings\textemdash and under ideal sliding window conditions\textemdash all models explored in this study perform comparably to sliding window, while providing significant speed improvements.

Future work includes extending this analysis to additional targets and detection models, notably the {\em you only look once (YOLO)} family of models \cite{redmon2015look, Redmon2016YOLO9000BF, Redmon2018YOLOv3AI}. The most recent incarnation (version 3) specifically addresses small, crowded object fields, which is of primary concern for small targets like cars in EO imagery. Additionally, for large objects like oil and gas fracking wells we intend to explore multi-scale (multi-resolution) approaches to object detection with the goal of maintaining performance while boosting area coverage rates. Lastly, we intend to explore detection accuracy and timing results for performing multi-class object detection, which presents additional challenges if target sizes vary drastically in size and in properties between the classes.

\section*{Acknowledgment}
The authors would like to thank Michael Harner, Tyler Kuhns, Andy Lam, Stephen O'Neill, and Frank Jones for useful discussions on the topics of electro-optical imagery, deep learning, and also for assisting with the manual curation of the oil and gas fracking well images.

\bibliographystyle{IEEEtran}
\bibliography{refs}

\begin{table*}[htb!]
    \large
    \centering
    \begin{tabular}{|c|c|c|c|c|c|}
    \hline
        {\bf Model Architecture} & {\bf CNN Backbone} & {\bf max $\mathrm{F_{1}}$} & {\bf AP} & {\bf FPS} & {\bf Cov. Rate (sq km/sec.)} \\
    \hline \hline
    RetinaNet & ResNet-50 & 0.872 & 0.893 & {\bf 7.1} & {\bf 17.06} \\\cline{2-6}
              & ResNet-101 & 0.878 & 0.905 & 5.9 & 14.17 \\\cline{2-6} 
              & ResNeXt-101 & {\bf 0.890} & {\bf 0.920} & 4.7 & 11.29 \\ 
    \hline \hline
    GHM & ResNet-50 & {\bf 0.884} & 0.912 & {\bf 6.9} & {\bf 16.58} \\\cline{2-6}
        & ResNet-101 & 0.878 & {\bf 0.916} & 5.8 & 13.93 \\\cline{2-6}
        & ResNeXt-101 & 0.871 & 0.905 & 4.4 & 10.57 \\ 
    \hline \hline
    Faster R-CNN & ResNet-50 & 0.875 & 0.890 & {\bf 6.7} & {\bf 16.10} \\\cline{2-6}
                 & ResNet-101 & 0.881 & 0.892 & 5.6 & 13.45 \\\cline{2-6} 
                 & ResNeXt-101 & {\bf 0.882} & {\bf 0.897} & 4.4 & 10.57 \\ 
    \hline \hline
    Grid R-CNN & ResNet-50 & 0.873 & {\bf 0.890} & {\bf 5.8} & {\bf 13.93} \\\cline{2-6}
              & ResNet-101 & 0.880 & 0.877 & 4.9 & 11.77 \\\cline{2-6} 
              & ResNeXt-101 & {\bf 0.881} & 0.880 & 5.0 & 12.01 \\ 
    \hline \hline
    Cascade R-CNN & ResNet-50 & 0.869 & 0.870 & {\bf 5.8} & {\bf 13.93} \\\cline{2-6}
                  & ResNet-101 & 0.864 & 0.873 & 5.0 & 12.01 \\\cline{2-6} 
                  & ResNeXt-101 & {\bf 0.883} & {\bf 0.884} & 4.0 & 9.61 \\ 
    \hline \hline
    Double-Head R-CNN & ResNet-50 & 0.879 & 0.891 & {\bf 4.3} & {\bf 10.33} \\\cline{2-6}
                      & ResNet-101 & {\bf 0.881} & {\bf 0.897} & 3.9 & 9.37 \\\cline{2-6} 
                      & ResNeXt-101 & 0.869 & 0.886 & 3.2 & 7.69 \\ 
    \hline
    \end{tabular}
    \caption{{\bf Oil and Gas Fracking Wells: }Performance and timing results.}
    \label{tab:well_table_results}
\end{table*}

\begin{table*}[htb!]
    \large
    \centering
    \begin{tabular}{|c|c|c|c|c|c|}
    \hline
        {\bf Model Architecture} & {\bf CNN Backbone} & {\bf max $\mathrm{F_{1}}$} & {\bf AP} & {\bf FPS} & {\bf Cov. Rate (sq km/sec.)} \\
    \hline \hline
    RetinaNet & ResNet-50 & {\bf 0.719} & {\bf 0.661} & {\bf 11.2} & {\bf 0.363} \\\cline{2-6}
              & ResNet-101 & 0.701 & 0.639 & 8.6 & 0.279 \\\cline{2-6} 
              & ResNeXt-101 & 0.662 & 0.602 & 5.9 & 0.191 \\ 
    \hline \hline
    GHM & ResNet-50 & 0.654 & 0.583 & \textbf{11.0} & \textbf{0.356} \\\cline{2-6}
        & ResNet-101 & \textbf{0.663} & \textbf{0.591} & 8.5 & 0.275 \\\cline{2-6}
        & ResNeXt-101 & 0.656 & 0.588 & 5.9 & 0.191 \\ 
    \hline \hline
    Faster R-CNN & ResNet-50 & 0.747 & 0.685 & \textbf{9.7} & \textbf{0.314} \\\cline{2-6}
                 & ResNet-101 & \textbf{0.752} & \textbf{0.691} & 7.8 & 0.253 \\\cline{2-6} 
                 & ResNeXt-101 & 0.747 & 0.684 & 5.5 & 0.178 \\ 
    \hline \hline
    Grid R-CNN & ResNet-50 & 0.756 & 0.697 & \textbf{7.6} & \textbf{0.246} \\\cline{2-6}
              & ResNet-101 & \textbf{0.758} & \textbf{0.699} & 6.4 & 0.207 \\\cline{2-6} 
              & ResNeXt-101 & 0.757 & \textbf{0.699} & 6.6 & 0.214 \\ 
    \hline \hline
    Cascade R-CNN & ResNet-50 & 0.751 & 0.687 & \textbf{7.9} & \textbf{0.256} \\\cline{2-6}
                  & ResNet-101 & \textbf{0.753} & \textbf{0.689} & 6.5 & 0.211 \\\cline{2-6} 
                  & ResNeXt-101 & 0.750 & 0.686 & 4.9 & 0.159 \\ 
    \hline \hline
    Double-Head R-CNN & ResNet-50 & 0.752 & 0.691 & \textbf{4.5} & \textbf{0.146} \\\cline{2-6}
                      & ResNet-101 & \textbf{0.754} & \textbf{0.695} & 4.0 & 0.130 \\\cline{2-6} 
                      & ResNeXt-101 & 0.752 & 0.693 & 3.3 & 0.107 \\ 
    \hline
    \end{tabular}
    \caption{{\bf Small Cars: }Performance and timing results.}
    \label{tab:car_table_results}
\end{table*}

\end{document}